# A Cuckoo Quantum Evolutionary Algorithm for the Graph Coloring Problem

Yongjian Xu and Yu Chen[0000-0002-8118-7262]

School of Science, Wuhan University of Technology, Wuhan, 430070, China.
ychen@whut.edu.cn

**Abstract.** Based on the framework of the quantum-inspired evolutionary algorithm, a cuckoo quantum evolutionary algorithm (CQEA) is proposed for solving the graph coloring problem (GCP). To reduce iterations for the search of the chromatic number, the initial quantum population is generated by random initialization assisted by inheritance. Moreover, improvement of global exploration is achieved by incorporating the cuckoo search strategy, and a local search operation, as well as a perturbance strategy, are developed to enhance its performance on GCPs. Numerical results demonstrate that CQEA operates with strong exploration and exploitation abilities, and is competitive to the compared state-of-the-art heuristic algorithms.

## 1    Introduction

Graph Coloring Problem (GCP) is one of the most important combinatorial problems in the scientific and engineering fields [1]. Due to its NP-completeness, heuristic algorithms have been widely investigated to develop efficient algorithms for large-scale GCPs. State-of-the-art heuristic algorithms for GCPs include the cuckoo search algorithm [2-5], the ant colony algorithm [6-7], the genetic algorithm [8-9], the memetic algorithm [10-11] and the tabu search algorithm [12] and so on. Recently, the marriage in honey bees optimization [13] and the DNA algorithm [14] are also developed for GCP.

Most of heuristic algorithms employ population-based search schemes, which enhance their convergence performance, but has the side-effect of heavy complexity. The quantum-inspired evolutionary algorithm (QEA) [15] operates with a small population, and thus, is of less complexity compared to most of population-based heuristic algorithms. Unfortunately, the small population would inevitably reduce its ability of global exploration, and sometimes results in local convergence to problems with complicated landscapes.

Imitating the behavior of cuckoos, the cuckoo search algorithm (CSA) with a few parameters has been successfully applied in engineering applications [16-18]. Introducing the quantum representation to the framework of CSA, Djelloul *et al*. [2] proposed a quantum inspired cuckoo search algorithm for graph coloring. However, falling into an optimal local trap and high runtime are two of the disadvantages of this research [1].

Recalling that QEA has low complexity arising from its small population size, we propose a novel cuckoo quantum evolutionary algorithm (CQEA) for the problem of graph coloring, where the cuckoo search (CS) schema is introduced to improve global exploration of QEA. Meanwhile, the inherited initialization strategy and special search strategies are developed to improve its performance on GCPs.

The rest of this article is organized as follows. Section 2 presents the formulation of GCP, and Section 3 elaborates details of CQEA for GCP. Performance of CQEA is validated in Section 4 by numerical experiments, and finally we conclude the work in Section 5.

## 2    Problem Description

In this paper, the GCP refers to the problem of vertex coloring. Given an undirected graph $G(V,E)$, where $V$ represents the set of vertices and $E$ is the set of edges, the GCP is dedicated to minimize the amount of colors to paint adjacent vertices by inconsistent colors. If $k$ different colors are sufficient to coloring adjacent vertices with different colors, $G(V,E)$ is called $k$-colorable. The smallest value of $k$ such that $G$ is called $k$-colorable is named as the *chromatic number* of $G(V,E)$.

If not confused, an undirected graph $G(V,E)$ is denoted as $G$. The GCP to get the *chromatic number* of $G$ can be model as a bi-level combinatorial optimization problem:

$$\begin{aligned} &\min k \\ s.t. \quad &\min f(S) = \sum_{(u,v) \in E} \delta_{uv} = 0, \\ &s.t. \begin{cases} \delta_{uv} = \begin{cases} 1, & if\ i = j, \\ 0, & if\ i \neq j, \end{cases} u \in V_i, v \in V_j; \\ S = (V_1, V_2, \cdots, V_k), \bigcup_{i=1}^{k} V_k = V, V_i \cap V_j = \varnothing, i \neq j; \\ i, j = 1, 2, \cdots, k. \end{cases} \end{aligned} \quad (1)$$

where $S = (V_1, V_2, \cdots, V_k)$ is a partition of $V$ that represents an assignment of colors of vertices. When $\delta_{uv} = 1$, incident vertices of edge $(u,v)$ are assigned the consistent color, and $(u,v)$ is named as a *conflicting edge*. Then, $f(S)$, the total amount of conflicting edge of $S$, is the objective function that evaluates qualities of color assignments. To get the *chromatic number* and the corresponding color assignment $S$, one must iteratively search the graph to minimize $f(S)$ for successively increasing/decreasing $k$, which is tedious and time consuming.

## 3  The Cuckoo Quantum Evolutionary Algorithm for GCPs

In this section, we first introduce the representation of the solution and outline the framework of the cuckoo quantum evolutionary algorithm. Then, details of population initialization, search strategy, cuckoo search operation, and perturbance strategy are presented.

### 3.1  Representation of the Solution to the Graph Coloring Problem

For an undirected simple graph $G$ of $n$ vertices, a color assignment with $k$ colors can be represented by a $k \times n$ binary matrix, where each column includes only one "1" bit [2]. Fig. 1 illustrates an example that a graph of five vertices is colored by three colors. For a graph included in Fig.1(a), the assignment of color illustrated in Fig.1(b) can be represented by the $3 \times 5$ binary matrix in Fig.1(c). When nodes 1 and 4 are colored by the first color (namely yellow), elements at the first row of columns 1 and 4 are set to be "1"; for nodes 2 and 3 painted by the second color (namely green), and node 5 given the third color (namely blue), we know the "1"-bits in columns 2, 3 and 5 are located at rows 2, 2 and 3, respectively.

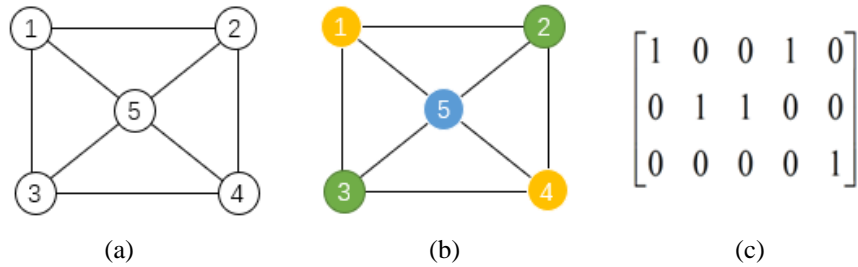

(a)                     (b)                     (c)

**Fig. 1.** (a) an undirected simple graph; (b) an assignment of color by 3 colors; (c) binary representation of color assignment.

Since a binary matrix $\mathbf{x}$ representing an assignment of color is equivalent to a partition $S$ of the vertex set $V$, its quality can be evaluated by the number of conflicting edges, that is,

$$f(\mathbf{x}) = f(S).  \qquad (2)$$

### 3.2  Quantum Matrix in the Cuckoo Quantum Evolutionary Algorithm

In the proposed cuckoo quantum evolutionary algorithm (CQEA), individuals are not binary matrices of coloring assignments but quantum matrices that present probability models about distributions of colors. An assignment of $k$ colors to $n$ vertices can be modelled by a $2k \times n$ quantum matrix

$$\mathbf{q} = \begin{bmatrix} \alpha_{11} & \alpha_{12} & \cdots & \alpha_{1N} \\ \beta_{11} & \beta_{12} & \cdots & \beta_{1N} \\ \vdots & \vdots & \vdots & \vdots \\ \alpha_{K1} & \alpha_{K2} & \cdots & \alpha_{KN} \\ \beta_{K1} & \beta_{K2} & \cdots & \beta_{KN} \end{bmatrix},$$

where $\alpha_{i,j}^2 + \beta_{i,j}^2 = 1, i = 1,\ldots,k, j = 1,\ldots,n$. $\alpha_{i,j}^2$ denotes the probability to assign the $i^{th}$ color to vertex $j$, and $\beta_{i,j}^2$ denotes the probability not to assign the $i^{th}$ color to vertex $j$. By performing the quantum measurement operation [15] on $\mathbf{q}$, a binary matrix $\mathbf{x}$ can be generated. Since there could be more than one "1" in a column of $\mathbf{x}$, a repair operation is necessary to transfer the generated $\mathbf{x}$ to a solution of GCP.

### 3.3 Framework of the Cuckoo Quantum Evolutionary Algorithm

The CQEA for GCP consists of two nested iterations. The inner loop tries to minimize number of conflicting edges for given color number $K$, and the external one is performed to minimize color numbers $K$ to get the chromatic number of the graph $G$. Termination criterions of loops are set as follows.

If iteration number of the external loop reaches a preset value or the chromatic number of $G$ is obtained, termination-condition 1 is satisfied; it quits the inner loop if the maximum number of generations is reached or a color assignment with no conflicting edges is obtained. Framework of the CQEA is presented as follows.

Framework of the CQEA
Begin
   $gen \leftarrow 0$;
  Initialize the color number $K$ and set solution archive $B(gen)=\varnothing$;
    while (termination-condition 1 is not satisfied) do
       $t \leftarrow 0$;
  Initialize the quantum population $Q(t)$ and the solution population $P(t)$;
      evaluate $P(t)$ and update the solution archive $B(gen)$;
      perform the local search on $B(gen)$;
      while (termination-condition 2 is not satisfied) do
          $t \leftarrow t+1$;
          generate $P(t)$ by $Q(t-1)$;
          perform the local search on $P(t)$;
          evaluate $P(t)$ and update $B(gen)$;

```
                Generate  Q(t)  by applying the cuckoo search operation to
Q(t−1) ;
                apply the perturbance strategy to update  Q(t) and  B(gen) ;
        end
    If  f(B(gen)) = 0
        K=K −1;
    end
            gen ← gen +1;
    end
end
```

Note that CQEA performs iterations on with its quantum population and the solutions to GCP are binary matrices. Thus, it generates binary matrices by implementing the quantum measurement operation on quantum matrices [15]. In the quantum measurement operation, an infeasible color assignment could be generated with uncolored nodes or nodes painted by more than one colors [2]. Then, a feasible color assignment is generated by the repair operation detailed as follows.

1. For a vertex with multiple colors, randomly choose one to color the vertex;
2. for a vertex that is not assigned with a color, randomly choose a color to paint the vertex.

After generation of the solution population, a binary matrix **x** that represents an assignment of color is evaluated by (2). While the solution archive is empty, the best solution with the least number of conflicting edges in $P(t)$ is put into $B(gen)$; if is not empty, the best solution in $P(t) \cup B(gen)$ is kept in $B(gen)$.

### 3.4    Initialization of the CQEA

**Initialization of color number $K$.**
Since the CQEA is implemented by successive iterations of the outer loop, promising initialization of $K$ is helpful to reduce the number of iterations to get the chromatic number. At the beginning, it sets $K=\Delta(G)+1$ according to the following theorem [19].

**Theorem 1.** Given undirected simple graph $G$, it holds $k \leq \Delta(G)+1$, where $\Delta(G)$ is the degree of $G$, and $k$ is the chromatic number of $G$.

**Initialization of the Quantum Population and the Solution Population.**
There are two different strategies employed to initialize the quantum population. While $gen = 0$, the quantum population is randomly initialized; when $gen > 0$, it is generated by the inheritance initialization strategy.

*Random Initialization.*

At the beginning, the state expressed by each individual in the population is the equal probability superposition of all its possible states [2, 15]. By setting $K = \Delta(G)+1$, each individual in the initial quantum population is taken as

$$\begin{bmatrix} \alpha_{11} & \alpha_{12} & \cdots & \alpha_{1N} \\ \beta_{11} & \beta_{12} & \cdots & \beta_{1N} \\ \vdots & \vdots & \vdots & \vdots \\ \alpha_{K1} & \alpha_{K2} & \cdots & \alpha_{KN} \\ \beta_{K1} & \beta_{K2} & \cdots & \beta_{KN} \end{bmatrix} = \begin{bmatrix} 1/\sqrt{2} & 1/\sqrt{2} & \cdots & 1/\sqrt{2} \\ 1/\sqrt{2} & 1/\sqrt{2} & \cdots & 1/\sqrt{2} \\ \vdots & \vdots & \vdots & \vdots \\ 1/\sqrt{2} & 1/\sqrt{2} & \cdots & 1/\sqrt{2} \\ 1/\sqrt{2} & 1/\sqrt{2} & \cdots & 1/\sqrt{2} \end{bmatrix}. \quad (3)$$

Once the quantum population $Q(0)$ is generated, the solution population $P(0)$ can be generated by perform quantum measurement on each individual of $Q(0)$.

*Inheritance Initialization.*

If CQEA has successfully colored $G$ by $K$ colors, it attempts to color $G$ by $K-1$ colors. To further improve searching efficiency of the upcoming iteration, the inheritance initialization is performed to generate an initial quantum population, which can be implemented by obtained best binary matrix $A$ and the corresponding quantum matrix $Q_A$ obtained with $K$ feasible colors. Detailed operations of the inheritance initialization are described as follows.

1. Take the best binary matrix $A$ and the corresponding quantum matrix $Q_A$ obtained by $K$ ($K \leq \Delta(G)+1$) colors. Locate zero rows[1] of the binary matrix $A$, and denote row indexes of zero rows as $n_j, j = 1,\ldots,row_0$.
2. If $row_0 > 0$, $G$ can be colored by $K - row_0$ colors. Set $K = K - row_0$. Delete zero rows of $A$ and the corresponding rows of $Q_A$, i.e., rows $2n_j - 1, 2n_j (j = 1,\ldots,row_0)$ of $Q_A$.
3. If $row_0 = 0$, locate a row of $A$ that has the least number of "1" and label it as row $n_0$. Since row $n_0$ has the least number of "1", we try to delete the corresponding color to get the initial coloring of iteration $g$. Then, delete row $n_0$ of $A$ and rows $2n_0 - 1, 2n_0$ of $Q_A$, where the number of colors is $K - 1$.

In this way, we can get an updated quantum matrix $Q_A$ and the corresponding binary matrix $A$. Take them as individuals in $Q(0)$ and $P(0)$, respectively. Other individuals of $Q(0)$ are generated by random initialization. Performing quantum measurement on $Q(0)$, one can get the other members of $P(0)$.

---

[1] Zero rows are rows of a matrix where all elements are equal to "0".

### 3.5 Local Search in the Solution Space

Inspired by the mutation operator proposed in [2] and the simple decentralized graph coloring (SDGC) algorithm proposed in [20], we introduce a local search performed in the solutions space to improve local exploitation of CQEA. Given a set $P$ of binary matrix, the local search is performed for every $\mathbf{x} \in P$.

1. randomly transverse vertices of graph $G$; according to the color assignment represented by $\mathbf{x}$, record the collection of alternative colors $AC_j$ for vertex $v_j$, $j = 1, 2, \ldots, n$;
2. generate a candidate solution $\mathbf{y}$ as follows. For vertex $v_j$ with nonempty $AC_j$, assign the smallest color index in $AC_j$ to $v_j$ if $P$ is the solution archive $B(gen)$; otherwise, randomly assign a color index in $AC_j$ to $v_j$;
3. replace $\mathbf{x}$ by $\mathbf{y}$ if $f(\mathbf{y}) < f(\mathbf{x})$.

Note that the local search applies different strategies when adjusting color indexes of vertices. While it is applied to the best solution set $B(gen)$, it is assigned the smallest available color index to attempt to reduce the amount of color number $K$; however, when searching neighborhoods of $P(t)$, a randomly selected color index is selected to try to improve global convergence by introducing randomness to the result.

The local search is repeated for *Limit* times before a solution without conflicting edges is generated.

### 3.6 The Cuckoo Search

Global exploration of CQEA is enhanced by introducing the cuckoo search (CS) to the quantum population [16-18]. For a quantum population $Q$, the cuckoo search to generate new quantum population $Q_{new}$ and associated update of $B(gen)$ are performed for each quantum matrix $\mathbf{q} \in Q$.

1. perform the Levy flight on $\mathbf{q}$ to generate new $\mathbf{q}_{new}$; generate binary matrix $\mathbf{x}$ and $\mathbf{x}_{new}$ by $\mathbf{q}$ and $\mathbf{q}_{new}$, respectively;
2. evaluate $\mathbf{q}$ and $\mathbf{q}_{new}$ by $f(\mathbf{q}) = f(\mathbf{x})$ and $f(\mathbf{q}_{new}) = f(\mathbf{x}_{new})$, respectively;
3. if $f(\mathbf{q}_{new}) < f(\mathbf{q})$, replace $\mathbf{q}$ and $\mathbf{x}$ by $\mathbf{q}_{new}$ and $\mathbf{x}_{new}$, respectively; otherwise, perform a random walk for $\mathbf{q}_{new}$ and generate a $\mathbf{x}_{new}$ to replace $\mathbf{q}$ and $\mathbf{x}$;
4. if $\mathbf{x}_{new}$ is better than a binary matrix $\mathbf{y} \in B(gen)$, replace $\mathbf{y}$ by $\mathbf{x}_{new}$.

### 3.7 The Perturbance Strategy

The quantum population $Q(t)$ and archive $B(new)$ are further updated by performing the perturbance strategy.

1. According to the color assignment represented by $B(new)$, traverse the graph to locate all conflict vertices of $G$;
2. Sort quantum matrices $\mathbf{q}$ in $Q(t)$ in ascending order of $f(\mathbf{q})$;
3. For all quantum matrices belonging to the second half of sorted $Q(t)$, resort elements corresponding to the conflicting vertices to $1/\sqrt{2}$;
4. For conflicting vertex $v$ of $G$, denote the collection of its adjacent vertices as $V_v$;
5. For $\mathbf{x} \in B(new)$, assign the same color to the non-adjacent vertices in $V_v$; Record available colors of $v$ by $AC_v$; If $AC_v \neq \varnothing$, randomly assign a color in $AC_v$ to $v$ when the color of $v$ is not in $AC_v$;

## 4   Experimental results

In order to evaluate the efficiency of the CQEA, it is implemented in MATLAB R2020a. The running environment is Intel(R) Core(TM) i7 CPU 860 @ 2.80GHz, system memory 8GB and Win7 operating system. Benchmark problems are selected from the graph coloring instance library2. Numerical results for 10 independent runs are collected in Table 1, where the obtained color number and iterations are recorded.

Table 1. Numerical Results of CQEA on Benchmark Graph Coloring Problems

| Problems | | | | Popula-tion Size | Number of Colors | | | | Ieterations |
|---|---|---|---|---|---|---|---|---|---|
| Instance | $|V|$ | $|E|$ | Best Known | | Min | Max | Mean | Std | Mean |
| myciel3 | 11 | 20 | 4 | 10 | 4 | 4 | 4 | 0 | 1 |
| myciel4 | 23 | 71 | 5 | 10 | 5 | 5 | 5 | 0 | 1 |
| queen5_5 | 25 | 160 | 5 | 10 | 5 | 5 | 5 | 0 | 2.8 |
| queen6_6 | 36 | 290 | 7 | 10 | 7 | 8 | 7.7 | 0.458 | 87.5 |
| myciel5 | 47 | 236 | 6 | 6 | 6 | 6 | 6 | 0 | 1 |
| huck | 74 | 301 | 11 | 6 | 11 | 11 | 11 | 0 | 1 |
| jean | 80 | 254 | 10 | 6 | 10 | 10 | 10 | 0 | 1 |
| david | 87 | 406 | 11 | 6 | 11 | 11 | 11 | 0 | 1.1 |
| games120 | 120 | 638 | 9 | 6 | 9 | 9 | 9 | 0 | 1 |
| miles250 | 128 | 387 | 8 | 6 | 8 | 8 | 8 | 0 | 8.7 |
| miles500 | 128 | 1170 | 20 | 6 | 20 | 21 | 20.5 | 0.5 | 1 |
| anna | 138 | 493 | 11 | 6 | 11 | 11 | 11 | 0 | 1 |
| fpsol2.i.1 | 496 | 11654 | 65 | 6 | 65 | 65 | 65 | 0 | 1 |

Numerical results in Table 1 indicate that the CQEA can successfully get the found the known optimal solution of 13 instances. Besides 11 problems that are optimized by CQEA with 100% success rate, the worst results of queen6_6 and miles500 are one greater than the known optimal results.

Competitiveness of CQEA is demonstrated by comparing it with state-of-the-art algorithms [3,9,2]. Table 2 tabulates the results on color number these algorithms. It is shown that CQEA and DBG [9] perform better on "queen6_6" than the other two. For the other 12 instances, all four algorithms can find the best known solutions.

---
2   http://mat.gsia.cmu.edu/COLOR/instances.html.

**Table 2.** Comparison between CQEA and state-of-the-art algorithms for GSPs

| Instance | k | | | |
|---|---|---|---|---|
| | CQEA | MCOA[3] | DBG[9] | QICSA[2] |
| myciel3 | 4 | 4 | 4 | 4 |
| myciel4 | 5 | 5 | 5 | 5 |
| queen5_5 | 5 | 5 | 5 | 5 |
| queen6_6 | **7** | 8 | 7 | 8 |
| myciel5 | 6 | 6 | 6 | 6 |
| huck | 11 | 11 | 11 | 11 |
| jean | 10 | 10 | 10 | 10 |
| david | 11 | 11 | 11 | 11 |
| games120 | 9 | 9 | 9 | 9 |
| miles250 | 8 | 8 | 8 | 8 |
| miles500 | 20 | 20 | 20 | 20 |
| anna | 11 | 11 | 11 | 11 |
| fpsol2.i.1 | 65 | 65 | 65 | 65 |

Since both CQEA and DBG [9] can find the currently known optimal solutions for 13 instances in Table 2, we compare their stability by success rates to get the best known number of colors. Results in Table 3 show that CQEA outperforms DBG [9] in terms of stability in 4 instances (namely david, miles250, anna, fpsol2.i.1), however, performs a bit worse than DBG on instance queen6_6. For miles120 and miles500, CQEA and DBG have a success rate of 100%. We conclude that CQEA is competitive to DBG, and more efficient than MCOA and QICSA.

**Table 3.** Comparison of success rate between CQEA and DBG [9]

| Algorithm | queen6_6 | david | games120 | miles250 | miles500 | anna | fpsol2.i.1 |
|---|---|---|---|---|---|---|---|
| CQEA | 4/15 | **15/15** | 15/15 | **15/15** | 10/15 | **15/15** | **15/15** |
| DBG[9] | 15/15 | 14/15 | 15/15 | 11/15 | 10/15 | 12/15 | 2/15 |

## 5  Conclusion

Considering that the quantum-inspired evolutionary algorithm can perform well with a small population, this paper develops a cuckoo quantum evolutionary algorithm (CQEA) to efficiently address the graph coloring problem. Enhancement of global exploration is achieved by introduction of general cuckoo search and specific strategies specially designed for the GCP, including an inherited initialization strategy, a local search strategy and a perturbance strategy. Numerical comparison based on benchmark problems demonstrates that the proposed CQEA can outperform most of the state-of-the-art algorithms, and is competitive to the DBG algorithm. The results also indicate that development of quantum-based heuristic algorithms could be a feasible solution to efficient address of large-scale GCPs. To further improve the performance of CQEA on large-scale GCPs, our future work would focus on designing efficient representation of solution and reduction strategy of graph scale.

# References

1. Mostafaie, T., Modarres, F., & Navimipour, N.J.: A systematic study on meta-heuristic approaches for solving the graph coloring problem. Computers and Operations Research, 120, 104850 (2020).
2. Djelloul, H., Layeb, A., & Chikhi, S.: Quantum inspired cuckoo search algorithm for graph colouring problem. International Journal of Bio-Inspired Computation, 7, 183-194 (2015).
3. Mahmoudi, S., & Lotfi, S.: Modified cuckoo optimization algorithm (MCOA) to solve graph coloring problem. Applied Soft Computing, 33, 48-64 (2015).
4. Aranha, C., Toda, K., & Kanoh, H.: Solving the Graph Coloring Problem Using Cuckoo Search. In: Tan Y., Takagi H., Shi Y. (eds) Advances in Swarm Intelligence. ICSI 2017. Lecture Notes in Computer Science, vol 10385, pp. 552-560. Springer, Cham (2017).
5. Zhou, Y., Zheng, H., Luo, Q., Wu, J., & Guangxi, N.: An improved Cuckoo Search Algorithm for Solving Planar Graph Coloring Problem. Applied Mathematics & Information Sciences, 7, 785-792 (2013).
6. Silva, A.F., Rodriguez, L.G., & Filho, J.F.: The improved Colour Ant algorithm: a hybrid algorithm for solving the graph colouring problem. International Journal of Bio-Inspired Computation, 16, 1-12 (2020).
7. Mohammadnejad, A., & Eshghi, K.: An efficient hybrid meta-heuristic ant system for minimum sum colouring problem. International Journal of Operational Research, 34, 269-284 (2019).
8. Marappan, R., & Sethumadhavan, G.: Solution to graph coloring problem using divide and conquer based genetic method. 2016 International Conference on Information Communication and Embedded Systems (ICICES), 1-5 (2016).
9. Douiri, S.M., & Elbernoussi, S.: Solving the graph coloring problem via hybrid genetic algorithms. Journal of King Saud University: Engineering Sciences, 27, 114-118 (2015).
10. Lü, Z., & Hao, J.: A memetic algorithm for graph coloring. European Journal of Operational Research, 203, 241-250 (2010).
11. Moalic, L., & Gondran, A.: Variations on memetic algorithms for graph coloring problems. Journal of Heuristics, 24, 1-24 (2018).
12. Hertz, A., & Werra, D.: Using tabu search techniques for graph coloring. Computing, 39, 345-351 (2005).
13. Bessedik, M., Toufik, B., & Drias, H.: How can bees colour graphs. International Journal of Bio-Inspired Computation, 3, 67-76 (2011).
14. Wang, Z., Wang, D., Bao, X., & Wu, T.: A parallel biological computing algorithm to solve the vertex coloring problem with polynomial time complexity. Journal of Intelligent & Fuzzy Systems, 40, 3957-3967 (2021).
15. Han, K., & Kim, J.: Quantum-inspired evolutionary algorithm for a class of combinatorial optimization. IEEE Transactions on Evolutionary Computation, 6, 580-593 (2002). doi: 10.1109/TEVC.2002.804320
16. Yang, X., & Deb, S.: Engineering optimisation by cuckoo search. International Journal of Mathematical Modelling and Numerical Optimisation, 1, 330-343 (2010).
17. Santillan, J., Tapucar, S., Manliguez, C., & Calag, V.: Cuckoo search via Lévy flights for the capacitated vehicle routing problem. Journal of Industrial Engineering International, 14, 293-304 (2018).
18. Yang, X.: Cuckoo search for inverse problems and simulated-driven shape optimization. Journal of Computational Methods in Sciences and Engineering, 12, 129-137 (2012).
19. West, D.B.: Introduction to Graph Theory, Second Edition. Prentice Hall, Upper Saddle River (2001).

20. Galán, S.F.: Simple decentralized graph coloring. Computational Optimization and Applications, 66, 163-185 (2017).